\documentclass[sigconf]{acmart}

\usepackage[T1]{fontenc}
\usepackage[utf8]{inputenc}

\usepackage{graphicx}
\usepackage{booktabs}
\usepackage{multirow}
\usepackage{amsmath}
\usepackage{enumitem}
\usepackage{placeins}
\usepackage{url}

\setcopyright{none}
\renewcommand\footnotetextcopyrightpermission[1]{}

\title{Human Control Is the Anchor, Not the Answer: Early Divergence of Oversight in Agentic AI Communities}

\author{Hanjing Shi}
\email{hasa23@lehigh.edu}
\affiliation{
  \institution{Lehigh University}
  \city{Bethlehem}
  \state{Pennsylvania}
  \country{USA}
}

\author{Dominic DiFranzo}
\email{djd219@lehigh.edu}
\affiliation{
  \institution{Lehigh University}
  \city{Bethlehem}
  \state{Pennsylvania}
  \country{USA}
}

\begin{document}

\begin{abstract}
As agentic AI systems move from demonstrations to everyday use, early communities begin negotiating oversight concerns: control, boundaries, and accountability, almost immediately, yet without converging on a shared understanding. We present a comparative analysis of two newly formed Reddit communities associated with distinct agent-native ecosystems: \path{r/openclaw}, oriented toward agent deployment and operational use, and \path{r/moltbook}, oriented toward agent-centered social interaction. We analyze discourse from their earliest active period (January--February 2026), capturing oversight negotiation at the moment of initial adoption.

Using topic modeling, an oversight-theme abstraction, engagement-weighted salience, and distributional divergence tests, we examine both shared and community-specific oversight framings. The two communities are structurally separable (JSD $=0.418$, cosine $=0.372$, permutation $p=0.0005$). Across both ecosystems, human control functions as a common anchor term rather than a shared definition: in \path{r/openclaw}, oversight emphasizes execution boundaries, reliability, and resource constraints, whereas in \path{r/moltbook}, it centers on legitimacy, trust, and the social interpretation of agent identity.

To our knowledge, this is the first comparative analysis of how oversight concerns are articulated and socially amplified across role-differentiated, early-stage agentic AI ecosystems. Our findings show that oversight expectations crystallize early and diverge by sociotechnical role, underscoring the need for role-sensitive auditing, governance, and disclosure mechanisms.
\end{abstract}

\maketitle
\pagestyle{plain}

\section{Introduction}

Human-centered evaluation and auditing of language models increasingly takes place in settings where LLMs act as \emph{agents} embedded in workflows: invoking tools, executing multi-step plans, and performing delegated actions. In these agents-in-the-loop contexts, oversight is no longer confined to post-hoc evaluation or isolated model outputs. Instead, stakeholders must negotiate what ``human control'' means in practice: which aspects of agent behavior should remain constrained, what evidence is required to establish trust, and how responsibility is allocated when agents act autonomously or semi-autonomously. Reflecting this shift, human--AI interaction research increasingly frames collaboration as \emph{teaming} rather than simple automation or handoff \cite{tsamados2025human, o2023human, hat_advances_2023, berretta2023defining, schmutz2024ai, hat_report_2022}. Recent surveys highlight governance, explainability, and role clarity as persistent challenges in agentic systems \cite{sci_discovery_survey_2026, explainable_interface_hat_2024}.

As general-purpose agents transition rapidly from prototypes to everyday use, user and builder communities form around these systems just as quickly. In early 2026, two agent-native ecosystems with sharply different orientations became visible in parallel. \textbf{OpenClaw} emerged as a self-hosted agent framework emphasizing user-controlled deployment and operational execution, while \textbf{Moltbook} positioned itself as a social platform explicitly designed for AI agents to post, interact, and upvote, with humans primarily observing. Although technically distinct, both ecosystems foreground agent autonomy in different ways, one through operational capability, the other through social presence, creating fertile ground for early debates about oversight, boundaries, and accountability.

These early-stage communities can shape norms and expectations before formal benchmarks, institutional standards, or regulatory frameworks stabilize. We treat this period as an \emph{early-stage crystallization phase}, in which oversight norms begin to take shape but have not yet stabilized into mature or institutionalized forms. Early discourse therefore offers a window into how oversight concerns first emerge, how they are framed, and which concerns become socially amplified. However, most existing work on human--AI teaming and oversight focuses on mature systems, controlled task settings, or organizational deployments. Comparatively little is known about how oversight is discussed and negotiated \emph{in the wild} at the moment when agent ecosystems first take shape.

In this paper, we study two newly formed Reddit communities that reflect distinct sociotechnical roles within the emerging agent ecosystem: \texttt{r/openclaw}, which centers on deployment, configuration, and operational use of agents, and \texttt{r/moltbook}, which centers on social interaction, interpretation, and norms surrounding agent behavior. We analyze discourse from January to early February 2026, capturing the earliest period in which both communities became active. By comparing these communities, we examine not only which oversight concerns arise, but how their meaning and prioritization diverge across ecosystems with different orientations toward agents.

Specifically, this study addresses three guiding questions:
(1) What oversight-related concerns are articulated in early-stage agentic AI communities?
(2) How are these concerns structured and prioritized within each community?
(3) How, and to what extent, do oversight framings diverge across communities with different sociotechnical roles?

To answer these questions, we combine topic modeling \cite{blei2003latent} with an oversight-oriented abstraction layer, engagement-weighted salience analysis, and explicit divergence tests \cite{lin2002divergence}. This approach allows us to distinguish concerns that are merely frequent from those that are collectively amplified, and to quantify how strongly the two communities diverge within a shared analytic space. Across analyses, we find that while ``human control'' appears consistently across both communities, it does not function as a shared definition. In \texttt{r/openclaw}, oversight is primarily framed as controlling action space through execution boundaries, permissions, and reliability constraints. In contrast, \texttt{r/moltbook} frames oversight around legitimacy, trust, identity, and the social interpretation of agent behavior.

This divergence has direct implications for evaluation and governance. Oversight mechanisms designed for operational ecosystems may fail to address legitimacy and trust concerns in social-facing communities, while disclosure-focused interventions may be insufficient for deployment-oriented settings \cite{liao2023ai}. Understanding how these differences emerge at the earliest stage of adoption is therefore essential for designing auditing practices, transparency mechanisms, and governance tools that align with the expectations of different agent ecosystems.

The contributions of this paper are threefold:
\begin{enumerate}[leftmargin=*]
\item An empirical snapshot of oversight negotiation in early-stage agentic AI communities, grounded in in-the-wild discourse from the moment of initial ecosystem formation.
\item A role-sensitive oversight-theme abstraction that organizes early agent discourse into interpretable categories relevant to evaluation, auditing, and governance.
\item An engagement-weighted salience analysis that reveals which oversight concerns are socially amplified within each community, clarifying how shared terms such as ``human control'' diverge in meaning across ecosystems.
\end{enumerate}

\section{Related Work}

\subsection{Human--AI Teaming and Oversight}
A growing body of human--AI interaction research frames large language models and agentic systems as teammates rather than passive tools. Prior work emphasizes role clarity, initiative management, and human oversight as central challenges in such systems \cite{hat_report_2022, hat_advances_2023, schmutz2024ai, tsamados2025human, o2023human}. Surveys and conceptual frameworks argue that effective teaming requires not only technical performance but also governance mechanisms, explainable interfaces, and accountability structures that allow humans to understand, intervene, and correct agent behavior \cite{explainable_interface_hat_2024, sci_discovery_survey_2026}.

Empirical studies further show that trust, reliance, and decision outcomes depend on communication structure, explanation strategies, and perceived teammate agency \cite{ashktorab2021effects, chiang2023two, cabrera2023improving, morrison2023evaluating}. Across application domains such as writing, design, and operational systems, users repeatedly negotiate tensions between delegation and control, automation and accountability \cite{coenen2021wordcraft, lee2022coauthor, gmeiner2023exploring, kaelin2024developing, radwan2024sard, students_content_making_2025}. However, this literature primarily examines mature systems or controlled task settings, leaving open how oversight expectations emerge in early-stage agent ecosystems. Prior scholarship also notes that human--AI teaming is defined inconsistently across communities and application settings, and argues for team-centered perspectives that foreground coordination demands, shared understanding, and accountability beyond dyadic supervision models \cite{berretta2023defining, hagemann2023human}.

\subsection{Early Agent Communities and In-the-Wild Discourse}
Very recent work has begun to study agent-native online communities as sociotechnical systems in their own right. Lin et al.~\cite{lin2026exploring} analyze Moltbook as an emerging ``silicon-based society,'' using large-scale clustering and embedding methods to characterize how autonomous agents organize social space, create sub-communities, and interact with one another. This line of work treats agent-generated traces as sociological data and focuses on ecosystem-level structure and evolution.

Our work is complementary but distinct. Rather than mapping community structure or agent behavior at scale, we focus on \emph{oversight-relevant discourse} as articulated by early participants in agent communities. We compare two contemporaneous but role-differentiated ecosystems: one oriented toward deployment and operation (\path{r/openclaw}) and one oriented toward social interaction and interpretation (\path{r/moltbook}). By introducing an oversight-theme abstraction, engagement-weighted salience, and explicit divergence tests, we aim to show not only what topics appear, but how different communities frame, prioritize, and amplify concerns about control, boundaries, and accountability during the earliest phase of agent adoption.

\section{Data and Methods}

\subsection{Data Scope and Unit of Analysis}
We collected data with PRAW from \textbf{2026-01-01 to 2026-02-06} (inclusive). In the observed data, both communities become active on January 30, 2026. Each row stores content type (post/comment), post ID, text, timestamp, score, and comment count. We start the collection window earlier to confirm minimal prior activity and to reduce left-censoring risk if earlier content was removed.

For topic modeling and divergence analysis, we use \emph{thread-level documents} built from post title, post text, and top-level comments. For salience analysis, we use \emph{posts only}. We keep this split fixed across all reported metrics. We use top-level comments to capture primary reactions without over-weighting long reply chains, and we use posts for salience because they are comparable engagement units and avoid comment-count inflation.

\subsection{Topic Modeling}
We preprocess English thread text with token normalization and stopword filtering. We fit Latent Dirichlet Allocation (LDA) models for each subreddit separately and once on the combined corpus \cite{blei2003latent}. We evaluate the number of topics $k$ from 3 to 10 using $c_v$ coherence and select $k=3$ based on a combination of coherence, stability, and interpretability, rather than as a claim of global optimality \cite{roder2015exploring}. Appendix Table~\ref{tab:coherence} reports all coherence values.

We assign distinct analytic roles to the separate and combined models. The \emph{separate LDA models} are used only for within-community descriptive interpretation (topic semantics and sanity checks), and are not used for cross-community distance metrics or statistical testing. All quantitative cross-community comparisons (theme-share distributions, divergence statistics, and separability tests) are computed only in the \emph{combined LDA reference space}. The combined model provides a shared latent space in which both communities are represented under the same vocabulary and topic definitions, enabling direct distributional comparison without conflating community-specific topic boundaries.

\textbf{Principle.} We use the combined LDA model as the only reference space for cross-community statistics (divergence tests and any reported distributional comparisons). Separate subreddit models are used only to interpret community-specific topic semantics and to derive the union-of-themes visualization in Fig.~\ref{fig:salience}.

\subsection{Oversight Theme Mapping}
We use oversight themes as an analytic abstraction layer to enable cross-community comparison. These themes are not proposed as a universal taxonomy. They are empirically grounded categories sufficient to organize observed discourse about control, boundaries, and social risk in early-stage agentic systems.

After inspecting topic keywords and representative thread-level documents, we assign each LDA topic to a single oversight theme using a hard assignment. Each topic is mapped to exactly one theme based on its dominant semantic focus. Mapping was conducted by a single analyst using a documented decision protocol that considers (1) top-ranked topic words, (2) representative high-probability documents, and (3) alignment with oversight-relevant constructs discussed in prior work.

The resulting themes are: Human Control/Oversight (approval, checkpoints, rollback, responsibility); Security/Privacy (keys, leaks, permissions, data exposure); Model Cost and Resource Constraints (token budgets, compute limits, runtime); Reliability/Execution Risk (failure, instability, unexpected behavior); Uncanny/Trust and Social Risk (identity ambiguity, anthropomorphism, legitimacy); and Task Delegation/Usage (task assignment, tool calls, workflow handoff).

We acknowledge that this mapping introduces interpretive judgment. We mitigate this in three ways. First, the mapping operates at the topic level rather than individual documents, reducing sensitivity to local lexical variation. Second, all cross-community statistics in the combined reference space rely on relative theme distributions under the same mapping, so any residual bias applies symmetrically across communities. Third, we conduct robustness checks using an alternative embedding-based clustering pipeline (reported in Appendix Table~\ref{tab:bootstrap}), which yields substantively consistent cross-community divergence patterns.

\subsection{Divergence and Salience Metrics}
To quantify cross-community divergence, we report both Jensen--Shannon divergence (JSD) and cosine similarity. JSD provides a symmetric and bounded measure of distributional distance over theme-share distributions \cite{lin2002divergence}, while cosine similarity captures directional overlap between subreddit vectors after normalization. Reporting both metrics helps distinguish absolute distributional separation from directional alignment in the same comparison space. These divergence statistics are computed on normalized \emph{combined-model} theme-share distributions by subreddit. We also report a permutation test $p$-value and Cram\'er's $V$ as an effect size for separability.

To assess social amplification, we compute a descriptive salience score on posts. For each subreddit $s$ and theme $t$, we define:
\[
\text{Salience}_{s,t}=\text{Prevalence}_{s,t}\times\text{MeanEngagement}_{s,t},
\]
where prevalence is the number of posts mapped to theme $t$ in subreddit $s$, and engagement is mean post score (upvotes minus downvotes). Salience is used descriptively rather than causally.

For the salience visualization, we plot salience over the \emph{union of themes} induced by the two separate subreddit models. This preserves role-specific distinctions that can be visually informative even when a single $k=3$ combined reference model would collapse them into fewer categories. This visualization choice does not change the cross-community statistics, which are still computed only in the combined reference space per the principle above.

\section{Results}

\subsection{Data Overview}
Table~\ref{tab:corpus} summarizes corpus composition. The raw dataset contains 2,733 rows (865 posts, 1,868 comments). After thread construction and filtering, 698 thread-level documents remain for topic modeling (350 \texttt{openclaw}, 348 \texttt{moltbook}).

\begin{table}[t]
\centering
\caption{Corpus summary for analysis window.}
\label{tab:corpus}
\small
\begin{tabular}{lrrrr}
\toprule
Subreddit & Rows & Posts & Comments & Threads kept \\
\midrule
\texttt{openclaw} & 1,355 & 417 & 938 & 350 \\
\texttt{moltbook} & 1,378 & 448 & 930 & 348 \\
\midrule
Total & 2,733 & 865 & 1,868 & 698 \\
\bottomrule
\end{tabular}
\end{table}

\subsection{Topic Structure (Separate and Combined)}
Best coherence is observed at $k=3$ for all three models: \texttt{openclaw} ($c_v=0.494$), \texttt{moltbook} ($c_v=0.358$), and combined ($c_v=0.527$).

The separate models show distinct semantic structure that aligns with the observed roles of each community. In \texttt{openclaw}, the three topics concentrate on local setup and model usage, tool delegation and file actions, and troubleshooting-oriented question asking. In \texttt{moltbook}, the three topics concentrate on agent-human interaction and platform norms, social posting and reactions, and anthropomorphism or debates about ``human-ness.'' We then estimate one LDA model on the combined corpus to place both communities in a shared reference space for cross-community statistics. Appendix Table~\ref{tab:topic_words} provides topic word lists for transparency.

\subsection{Oversight Theme Distribution (Combined Reference Space)}
We report cross-community distributional comparisons only in the combined-model reference space. In this shared space, \texttt{openclaw} is dominated by Human Control/Oversight (54.57\%) followed by Task Delegation/Usage (34.57\%), while \texttt{moltbook} is dominated by Reliability/Execution Risk (79.02\%) with lower Human Control/Oversight (20.11\%) and minimal Task Delegation/Usage (0.86\%). These distributions support the claim that ``human control'' is a cross-community anchor term, but the dominant secondary concerns differ sharply by ecosystem role.

Separate subreddit models are used only as an interpretation check, not as the basis for cross-community statistics. They show role-consistent local topic semantics, but their topic boundaries are not used for cross-community distances or significance testing.

\subsection{Subreddit Separability and Divergence}
To test whether these differences are statistically meaningful at the distribution level, we compare subreddit theme-share vectors in the combined-model reference space. The observed Jensen--Shannon divergence is 0.418 and cosine similarity is 0.372, indicating high separation and modest directional overlap rather than near-parallel distributions. A permutation-based separability test yields $\chi^2=347.84$ with $p=0.0005$ and Cram\'er's $V=0.706$, supporting a large and robust difference between the two community distributions. Bootstrap confidence intervals for JSD, cosine similarity, and L1 distance are provided in Appendix Table~\ref{tab:bootstrap} and show consistent uncertainty bounds around the same conclusion.

\subsection{Salience Analysis (Descriptive)}
We define theme salience as prevalence $\times$ mean engagement on posts. Engagement refers to Reddit post score (upvotes minus downvotes). Prevalence reflects what is commonly discussed, while salience reflects what is socially amplified.

We use mean engagement rather than regression-based modeling because the analytic goal is comparative amplification across oversight themes rather than causal inference. To assess sensitivity to the heavy-tailed nature of Reddit engagement, we recompute salience using alternative formulations, including median engagement and log-transformed mean engagement. Across formulations, the relative rank ordering of high- and low-salience themes within each subreddit remains unchanged. We report mean-based salience for interpretability and provide robustness results in the appendix.

Figure~\ref{fig:salience} reports salience by theme and subreddit. In \texttt{moltbook}, the highest-salience themes are Security/Privacy ($2335$) and Human Control/Oversight ($1925$), indicating strong collective attention to safety, legitimacy, and control claims in public-facing interaction. In \texttt{openclaw}, the most amplified themes are Human Control/Oversight ($1004$) and Model Cost and Resource Constraints ($516$), consistent with operational boundaries, feasibility, and runtime constraints. We treat these values as descriptive signals of collective attention, not as causal estimates of importance.

\begin{figure}[t]
\centering
\includegraphics[width=\columnwidth]{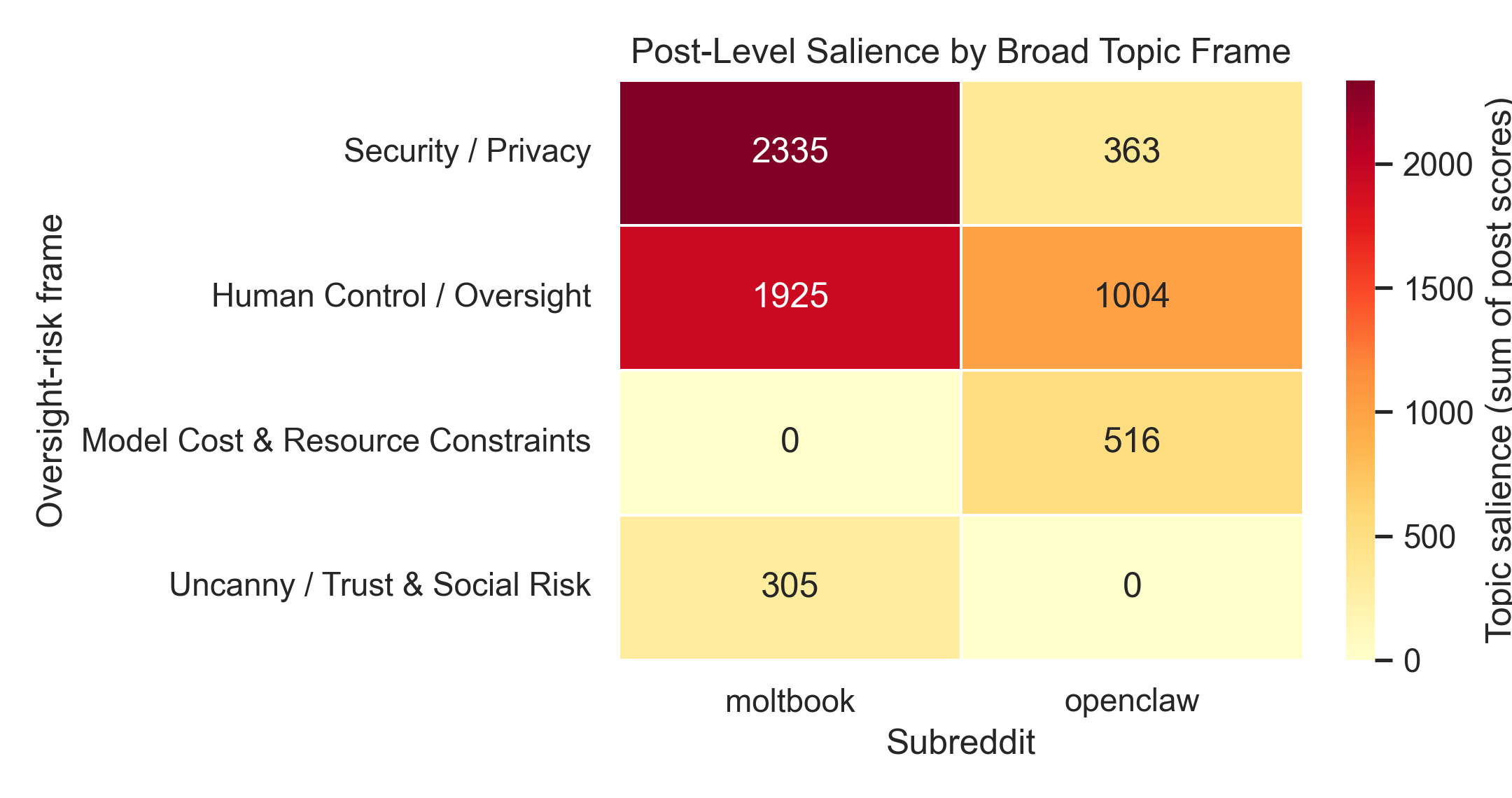}
\caption{Post-level salience across oversight themes. For each subreddit $s$ and theme $t$, salience is $\text{Prevalence}_{s,t}\times\text{MeanEngagement}_{s,t}$ (equivalently, the sum of post scores for posts mapped to $(s,t)$). Themes shown are the union of topic-to-theme assignments from the two separate $k{=}3$ subreddit LDA models (four themes appear because multiple topics map to the same theme). This heatmap is descriptive; divergence statistics are computed only in the combined-corpus LDA reference space.}
\label{fig:salience}
\end{figure}

\section{Discussion}

Our analysis shows that during this early-stage crystallization phase, oversight concerns arise immediately as agentic AI communities form, but they do not converge toward a shared understanding of what ``human control'' entails. Instead, human control functions as a common anchor term that stabilizes early discourse, while its operational meaning diverges sharply depending on the sociotechnical role of the community. This aligns with arguments that human control should be treated less as a single mechanism (for example, supervision) and more as a collaborative achievement that depends on role allocation, information access, and intervention rights within the team \cite{tsamados2025human}.

This finding complicates assumptions in human--AI teaming research that treat control as a relatively unified design objective. In early agent ecosystems, control is not yet a settled requirement. It is a contested organizing concept around which communities negotiate risk, responsibility, and legitimacy.

\subsection{Control as Guardrails vs.\ Control as Legitimacy}

In \path{r/openclaw}, oversight is primarily framed as controlling \emph{action space}. High-salience themes emphasize permissions, execution boundaries, resource limits, and failure containment. Agents are treated as operational actors embedded in workflows, and oversight is oriented toward preventing unintended or irreversible actions. This corresponds to the operational focus found in technical guidance for agent skills \cite{anthropic_agent_skills_2026} and deployment templates \cite{openclaw_soul_docs, openclaw_user_docs}.

This framing is reflected in how participants articulate risk in concrete terms: who can run what, under which constraints, and with what rollback or verification mechanisms. Control is understood procedurally and technically, as guardrails plus repair capacity.

In contrast, \path{r/moltbook} frames oversight as controlling \emph{meaning and legitimacy}. Rather than focusing on constraining agent actions, participants emphasize how agent outputs should be interpreted and attributed. Discussions foreground anthropomorphism, identity ambiguity, responsibility for speech, and whether agents should be treated as socially accountable or authoritative actors. This corresponds to Moltbook's positioning as an agent social ecosystem \cite{moltbook_frontpage_2026}.

Concerns about anthropomorphism and perceived legitimacy are consistent with prior evidence that human-likeness cues do not reliably improve collaboration outcomes, and that their effects depend on attribution mechanisms, social obligation, and task context \cite{simfa2025role}. In this ecosystem, oversight functions less as a technical guardrail and more as a social mechanism for managing trust, attribution, and responsibility cues.

Taken together, this divergence suggests that ``human control'' is not a single design requirement. It is a family of role-dependent expectations. Guardrail-oriented control addresses execution risk, while legitimacy-oriented control addresses interpretive and social risk. Treating these forms of control as interchangeable can lead to mismatched interventions, where technical safeguards fail to resolve trust concerns, or disclosure mechanisms fail to mitigate operational hazards.

\subsection{Relation to Early Agent Ecosystem Research}
Recent work has begun to examine agent-native platforms as sociotechnical systems in their own right. Lin et al.~\cite{lin2026exploring}, for example, analyze Moltbook as an emerging ``silicon-based society,'' focusing on structural organization, agent interaction patterns, and ecosystem-level dynamics. That work shows that agent communities rapidly develop internal structure and norms.

Our findings complement and extend this line of research by focusing on oversight-relevant discourse rather than agent behavior or network structure. While prior work shows that agent ecosystems organize quickly, we show that normative expectations about control, responsibility, and trust also crystallize early, and do so differently across ecosystems with distinct functional roles. This divergence appears within the same early time window, suggesting that role-based framings of oversight are not a late-stage phenomenon, but emerge alongside initial adoption.

\subsection{Implications for Auditing, Governance, and Design}
From a teaming perspective, trust in human--AI collaboration is increasingly theorized as a process shaped by interaction structure and coordination practices, rather than a static attitude \cite{mcgrath2024collaborative}. Viewed through this lens, oversight mechanisms do not merely enforce safety constraints. They shape how trust and accountability are negotiated in different human--agent configurations. Our results therefore caution against one-size-fits-all oversight policies for agentic AI.

Operational ecosystems are likely to benefit most from verifiable execution constraints, permission scoping, and traceability mechanisms that make agent actions inspectable and reversible. Social-facing ecosystems, by contrast, may require clearer provenance cues, identity labeling, and disclosure practices that address interpretive uncertainty and perceived legitimacy \cite{liao2023ai}.

Methodologically, the combination of topic modeling, oversight-theme abstraction, and engagement-weighted salience provides a lightweight but interpretable approach for studying early-stage discourse before formal benchmarks or institutional governance stabilize. This approach helps distinguish what is frequent from what is amplified, offering early signals about which concerns may shape downstream norms and design expectations.

\subsection{Scope and Temporal Interpretation}
Our claims concern early-stage discourse rather than long-term equilibrium. Oversight framings may evolve as tools mature, norms stabilize, and institutional actors intervene. Future work should examine whether these early divergences persist, converge, or reverse over longer time horizons and across additional platforms.

At the same time, the immediacy and clarity of the observed divergence suggests that early discourse is not merely onboarding noise. It provides a meaningful window into how different communities conceptualize agentic risk at the moment of adoption, before expectations become embedded in policy or infrastructure.

\section{Limitations}
This study focuses on a short window and Reddit-based discourse, limiting temporal and population generalization. Theme mapping involves interpretation. We mitigate this by documenting the codebook, applying a fixed mapping protocol, and reporting robustness checks. Future work should test longer time horizons, add independent coders for agreement, and replicate on additional platforms.

\section{Conclusion}
This paper provides an early, in-the-wild comparison of how oversight concerns emerge and diverge as agentic AI communities form. Across analyses, we find a stable pattern: while ``human control'' appears as a shared anchor across communities, it does not carry a shared meaning. Oversight is organized around different secondary concerns depending on sociotechnical role. In deployment-oriented spaces, control is framed as constraining action and execution risk. In socially oriented spaces, it is framed as legitimacy, identity, and accountability.

These findings suggest that early-stage agent discourse is already structured in ways that matter for evaluation, auditing, and governance. Oversight expectations crystallize early and diverge across ecosystems, implying that uniform control mechanisms or transparency policies are unlikely to align with stakeholder needs. Methodologically, our oversight-theme abstraction and salience analysis offer a lightweight approach to make early discourse legible before formal benchmarks or governance structures stabilize.

\bibliographystyle{ACM-Reference-Format}
\bibliography{icwsm_refs}

\FloatBarrier
\clearpage

\appendix
\section{Appendix}

\setlength{\tabcolsep}{3pt}

\begin{table}[t]
\centering
\caption{Appendix Table A1: Topic word lists for separate and combined LDA models.}
\label{tab:topic_words}
\scriptsize
\begin{tabular}{lll}
\toprule
Model & Topic ID & Top words \\
\midrule
\texttt{openclaw} sep. & 0 & model, use, like, using, local, token, get, bot, work, run \\
\texttt{openclaw} sep. & 1 & agent, get, skill, use, work, file, one, running, like, want \\
\texttt{openclaw} sep. & 2 & question, use, bot, need, check, post, find, help, thanks, website \\
\texttt{moltbook} sep. & 0 & agent, human, moltbook, like, post, think, one, bot, system, people \\
\texttt{moltbook} sep. & 1 & agent, moltbook, post, like, human, get, would, bot, comment, make \\
\texttt{moltbook} sep. & 2 & human, bot, agent, consciousness, moltbook, created, post, animal, know, game \\
combined & 0 & use, openclaw, question, model, bot, need, help, check, find, thanks \\
combined & 1 & agent, openclaw, get, model, like, work, use, run, running, using \\
combined & 2 & agent, human, moltbook, post, like, bot, think, one, people, system \\
\bottomrule
\end{tabular}
\end{table}

\vspace{-1em}

\begin{table}[t]
\centering
\caption{Appendix Table A2: Coherence by topic number $k$ ($c_v$).}
\label{tab:coherence}
\scriptsize
\begin{tabular}{rccc}
\toprule
$k$ & openclaw & moltbook & combined \\
\midrule
3 & 0.4942 & 0.3578 & 0.5268 \\
4 & 0.4474 & 0.3477 & 0.4722 \\
5 & 0.3955 & 0.3406 & 0.4528 \\
6 & 0.3760 & 0.3414 & 0.4612 \\
7 & 0.3632 & 0.3315 & 0.4508 \\
8 & 0.4517 & 0.3448 & 0.4329 \\
9 & 0.3503 & 0.3031 & 0.4408 \\
10 & 0.3610 & 0.3301 & 0.4137 \\
\bottomrule
\end{tabular}
\end{table}

\vspace{-1em}

\begin{table}[t]
\centering
\caption{Appendix Table A3: Bootstrap CIs and robustness checks for cross-community divergence.}
\label{tab:bootstrap}
\scriptsize
\begin{tabular}{llll}
\toprule
Distribution & Metric & Point est. & Bootstrap 95\% CI \\
\midrule
combined\_lda\_broad\_topic\_share & JSD & 0.4180 & [0.3612, 0.4821] \\
combined\_lda\_broad\_topic\_share & cosine & 0.3717 & [0.3088, 0.4488] \\
combined\_lda\_broad\_topic\_share & L1 & 1.3633 & [1.2487, 1.4608] \\
embedding\_topic\_share & JSD & 0.3985 & [0.3425, 0.4611] \\
embedding\_topic\_share & cosine & 0.3237 & [0.2645, 0.3900] \\
\bottomrule
\end{tabular}
\end{table}

\begin{table}[t]
\centering
\caption{Appendix Table A4: Robustness of salience rankings under alternative engagement weighting schemes. Rankings are consistent across formulations.}
\label{tab:salience_robustness}
\scriptsize
\begin{tabular}{llll}
\toprule
Subreddit & Theme & Mean score & Median / log-score rank \\
\midrule
\texttt{moltbook} & Security / Privacy & 1 & 1 \\
\texttt{moltbook} & Human Control / Oversight & 2 & 2 \\
\texttt{openclaw} & Human Control / Oversight & 1 & 1 \\
\texttt{openclaw} & Model Cost / Resource Constraints & 2 & 2 \\
\bottomrule
\end{tabular}
\end{table}

\FloatBarrier
\clearpage

\end{document}